\documentclass{article} 
\usepackage{iclr2018_conference,times}
\usepackage{hyperref}
\usepackage{url}
\usepackage{graphicx}
\usepackage{pbox}
\usepackage{wrapfig}
\usepackage{multirow}
\usepackage{float}
\graphicspath{{./figs/}}
\usepackage{cleveref}
\crefformat{footnote}{#2\footnotemark[#1]#3}

\title{
Fast Reading Comprehension with ConvNets}

\author{Felix Wu\thanks{A majority of the work was done while the author was interning at Google.}
\\
Department of Computer Science\\
Cornell University\\
Ithaca, NY, USA\\
\texttt{fw245@cornell.edu}
\AND 
Ni Lao, John Blitzer\\
Google Inc.\\
Mountain View, CA, USA\\
\texttt{\{nlao,blitzer\}@google.com}
\AND 
Guandao Yang, Kilian Q. Weinberger\\
Cornell University\\
Ithaca, NY, USA\\
\texttt{\{gy46,kqw4\}@cornell.edu}
}

\newcommand{\modelname}{Gated Linear Dilated Residual Network }
\newcommand{\modelshort}{GLDR }
\newcommand{\modelshorts}{GLDRs }

\newcommand{\specialcell}[2][c]{%
    \begin{tabular}[#1]{@{}c@{}}#2\end{tabular}}
\renewcommand{\paragraph}[1]{\vspace{-0.5ex}\textbf{#1}}

\newcommand{\nl}[1]{}
\newcommand{\fw}[1]{}
\newcommand{\jb}[1]{}
\newcommand{\kqw}[1]{}

\begin{document}

\maketitle

\begin{abstract}
State-of-the-art deep reading comprehension models are dominated by recurrent neural nets. Their sequential nature is a natural fit for language, but it also precludes parallelization within an instances and often becomes the bottleneck for deploying such models to latency critical scenarios. This is particularly problematic for longer texts.  
Here we present a convolutional architecture as an alternative to these recurrent architectures.  
Using simple dilated convolutional units in place of recurrent ones, we achieve results comparable to the state of the art on two question answering tasks, while at the same time achieving up to two orders of magnitude speedups for question answering.
\end{abstract}

\section{Introduction}
Recurrent neural networks such as LSTMs~\citep{hochreiter1997long} and GRUs~\citep{cho2014gru} are been very successful at simplifying certain natural language processing (NLP) systems, such as language modeling and machine translation.
The dominant of \emph{deep text understanding} models \citep{rajpurkar2016squad, joshi2017trivia, seo2016bidirectional} typically relies on recurrent networks to produce initial representations for the question and the document, and then apply attention mechanisms~\citep{bahdanau2014neural} to allow information passes between the two representations. The recurrent units are powerful structures capable of modeling complex long range interactions. However, their sequential nature precludes parallelization within training examples, and often become the bottleneck for deploying models to latency critical NLP applications. 
High latency is  especially critical for interactive question answering (for example as part of search engines or mobile assistants), as it requires the user to wait patiently for the answer.  

Recent development of ``attention only'' deep text models \cite{parikh2016attention, vaswani2017attention} in various tasks allows modeling of long range dependencies without regard to their distance. By parallelization within one instance, these models can have much better inference time than those which depend on recurrent units. However, their token-pair based attention requires $O(n^2)$ memory consumption within the GPUs, where $n$ denotes the length of the document. This quadratic growth prevents their use with most real-world documents, such as e.g. Wikipedia pages (e.g., \autoref{fig:memory} compares the memory usage of different types of models).

In this work we propose, \modelname (\modelshort), a different architecture to avoid recurrent units in text precessing.
More specifically, we use a combination of residual networks~\citep{he16residual}, dilated convolutions~\citep{yu2016dilated} and gated linear units~\citep{dauphin2016glu}. 

\subsection{Reading Comprehension Tasks}
Reading comprehension tasks focus on one's ability to read a piece of text and subsequently answer questions about it (see TriviaQA examples in Figure~\ref{txt:trivia_qa_samples}).
We follow the typical reading compression setting and assume that the correct answer can be given as a snippet of the original text. This reduces the problem to a search problem, where the question functions as a query. 

\begin{figure}[t]
  \centering
  \fbox{
  \begin{minipage}{0.8\textwidth}
	\textit{Q: Plato and Xenophon were both pupils of which Greek philosopher? }\\
	\textbf{Socrates} ... philosophy. He is an enigmatic figure known chiefly through the accounts of classical writers, especially the writings of his students Plato and Xenophon and the plays of his contemporary Aristophanes. ...
	\\\\
    \textit{Q: What is the next in the series: Carboniferous, Permian, Triassic, Jurassic?}\\
... The Jurassic North Atlantic Ocean was relatively narrow , while the South Atlantic did not open until the following \textbf{Cretaceous} period , when ...
	\\\\
	\textit{Q: Who was the choreographer of the dance troupe Hot Gossip?} \\
	\textbf{Arlene Phillips}, ... Lee Mack. Hot Gossip In Britain, Phillips first became a household name as the director and choreographer of Hot Gossip, a British dance troupe which she formed in 1974 ...
  \end{minipage}}
  \caption{Samples from TriviaQA~\citep{joshi2017trivia}}
  \label{txt:trivia_qa_samples}
\end{figure}

Sometimes these tasks simply involve answer type and query term matching, but they sometimes may contain discourse phenomena like coreference (e.g. \emph{What philosopher taught Plato and Aristophanes?} and \emph{Who was the choreographer of the dance troupe Hot Gossip?}) or even real world knowledge (e.g., answer \emph{What is the next in the series: Carboniferous, Permian, Triassic, Jurassic?} potentially involves understanding the semantics of \emph{next} and \emph{following}). Figure~\ref{txt:hard_example} shows an adversarial example of a question that is answered incorrectly by matching the first occurrence of the query word ``composed'' in the answer text.  
This study will use two popular reading comprehension tasks -- Trivia QA~\citep{joshi2017trivia}, and SQUAD~\citep{rajpurkar2016squad} -- as its test bed. Both tasks have openly available training and validation data  sets and are associated with competitions over a hidden test set on a public leaderboard. 

\begin{figure}[b]
  \centering
  \fbox{
  \begin{minipage}{0.8\textwidth}
	\textit{Q: Who composed the works The Fountains of Rome and The Pines of Rome in 1916 and 1924 respectively?} \\
	Adversary example . {\color{red}Wolfgang Amadeus Mozart} composed The Magic Flute , and Requiem . \textbf{Ottorino Respighi} ( ; 9 July 1879 - 18 April 1936 ) was an Italian violinist , composer and musicologist , best known for his three orchestral tone poems Fountains of Rome ( 1916 ) , Pines of Rome ( 1924 ) , and Roman Festivals ( 1928 ). 
  \end{minipage}}
  \caption{DrQA gives a wrong answer "Wolfgang Amadeus Mozart" rather than "Ottorino Respighi" by simply matching the verb "composed".}
  \label{txt:hard_example}
\end{figure}

Because of the sequential nature of documents and text, and 
complex long-distance relationships between words, recurrent neural networks (especially LSTMs~\cite{hochreiter1997long} and GRUs~\cite{cho2014gru}) are a natural class for modeling reading comprehension.  
Indeed, on both reading comprehension tasks we study here, every published result on the leader-board~\footnote{ \url{https://competitions.codalab.org/competitions/17208}}~\footnote{ \url{https://rajpurkar.github.io/SQuAD-explorer}} 
uses some kind of recurrent mechanism.
Below, we will discuss two specific models in detail, but we begin by motivating a one-dimensional convolution architecture 
for the reading comprehension task.

\subsection{Text Understanding with Dilated Convolutions}
\begin{figure}
	\centering
    \includegraphics[width=0.8\textwidth]{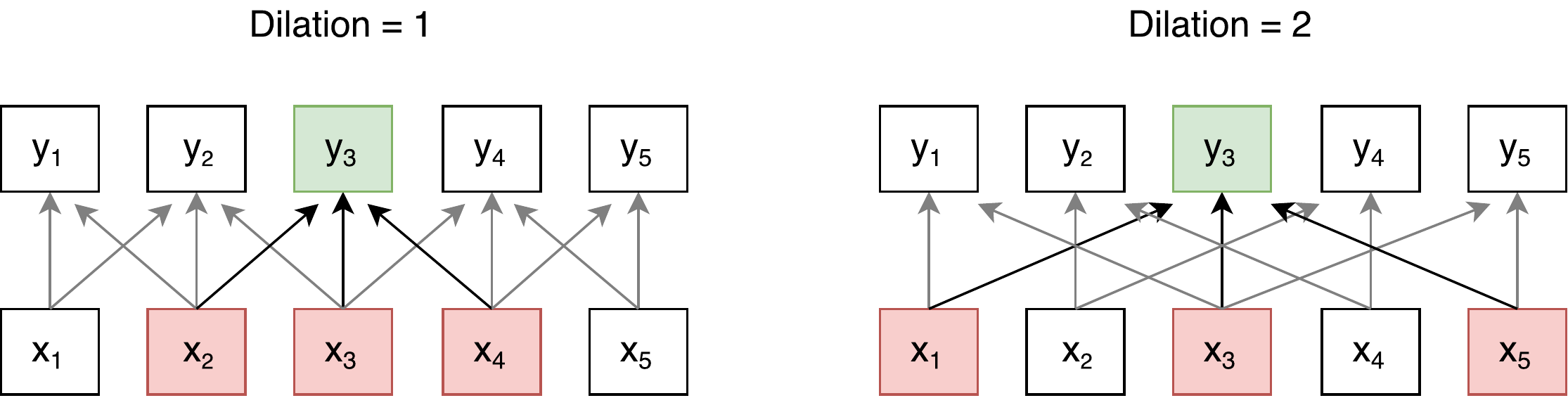}
    \caption{An illustration of dilated convolution. With a dilation of 1 (\emph{left}), dilated convolution reverts to standard convolution. With a dilation of 2 (\emph{right}) every other word is skipped, allowing the output $y_3$ to relate words $x_1$ and $x_5$ despite their large distance and a relatively small convolutional kernel. }
    \label{fig:dilation}
\end{figure}

Bidirectional recurrent units can in theory model arbitrarily long dependencies in text, but in practice we may be able to capture these dependencies through other mechanisms.  
We propose to substitute complicated and costly sequential models through simple feed-forward network architectures. 
There are two important criteria of language that LSTMs model, that we also want to capture.  First, we may need to model relationships between individual words, even 
when they are separated from each through many words 
(e.g. Figure~\ref{txt:trivia_qa_samples}).  
Second, we want to model the compositional nature of natural language semantics, where the meaning of large phrases are composed of the meaning of their sub-phrases.

These constraints lead us to choose dilated convolutional networks~\citep{yu2016dilated} with gated linear units~\citep{dauphin2016glu}.  
By increasing the receptive field  in our convolutional units, 
dilation can help to
model arbitrarily long-distance dependencies.  Unfortunately, the receptive region is pre-determined, which prevents us from examining long range dependencies in detail.  For instance, in the co-reference examples from Figure~\ref{txt:trivia_qa_samples}, we would need to directly convolve representations for ``his'' and ``Socrates'', but an increasing dilation will miss this.  
In practice, ``Socrates'' is combined with its context to give a fixed size representation for a long context.    
We alleviate some of this effect by using Gated Linear Units~\citep{dauphin2016glu} in our convolutions.  These units allow us to selectively retain (and compute gradients for) important features of low-level words and phrases, even at convolutions with larger dilations.  

\paragraph{Dilated Convolution.}
Given a 1-D convolutional kernel $\mathbf{k} = \left[ k_{-l}, k_{-l+1}, ...,  k_{l} \right]$ of size $2l+1$ and the input sequence $\mathbf{x} = \left[ x_1, x_2, ...,  x_n \right]$ of length $n$, a $d$ dilated convolution of $\mathbf{x}$ with respect the kernel $\mathbf{k}$ can be described as
\[
\left( \mathbf{k} * \mathbf{x} \right)_t = \sum_{i=-l}^{l} k_i \cdot x_{t - d \cdot i}
\] 
\begin{wrapfigure}{r}{0.2\textwidth}
\vspace{-2ex}
 \includegraphics[width=0.2\textwidth]{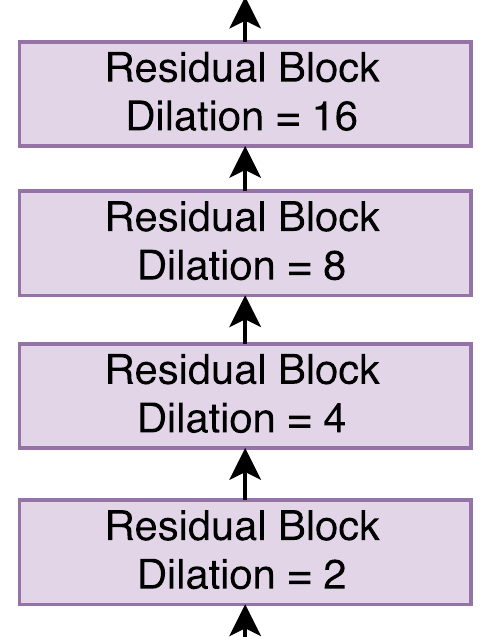}
 \vspace{-3ex}
 \caption{\small The receptive field of repeated dilated convolution grows exponentially with network depth.}\vspace{-2ex}
 \label{fig:block1}
\end{wrapfigure}
where $t \in \{1, 2, \cdots, n\}$. Here we assume zero-padding, so tokens outside  the sequence will be treated as zeros. 
Unlike normal convolutions (i.e. $d = 1$) that convolve each contiguous subsequence of the input sequence with the kernel, dilated convolution uses every $d^{th}$ element in the sequence, but shifting the input by one at a time.
\autoref{fig:dilation} shows an example of dilated convolution. Here, the green output is a weighted combination of the red input words. 

\begin{table}[t]
\centering 
\label{computations}
\begin{tabular}{l|cccc}
Layer type          & \specialcell{Computations\\per layer} & \specialcell{Minimum depth $D$\\to cover length $n$} & \specialcell{Longest\\computation path} & \specialcell{Overall\\computations} \\ \hline
Recurrent Units     & $O(w^2n)$      & $O(1)$        & $O({\color{red}n}D)$  & $O(w^2n)$  \\
Self-Attention      & $O(wn^2)$      & $O(1)$        & $O(D)$    & $O(w{\color{red}n^2})$              \\
Dilated Convolution & $O(kw^2n)$     & $O(\log(n))$   & $O(D)$    & \specialcell{$O(kw^2n D)$ or\\$O(kw^2n \log(n))$}            
\end{tabular}
\caption{Comparison among three sequence encoding layers with input sequence length $n$, network width $w$, kernel size $k$, and network depth $D$. Recurrent units and self-attention become slow as $n$ grows. When the receptive field of a dilated convolution covers the longest possible sequence $n$, its overall computation is proportional to $O(\log{n})$.}
\label{table:complexity}
\end{table}

\paragraph{Why Dilated convolution?} Repeated dilated convolution~\citep{yu2016dilated} increases the receptive region of ConvNet outputs exponentially with respect to the network depth, which results in drastically shortened computation paths. See Figure~\ref{fig:block1} for an illustration of an architecture with four dilated convolutional layers with exponentially increasing dilations. \autoref{table:complexity} shows a brief comparison between bidirectional recurrent units, self-attention, and dilated convolution. Self-attention suffers from the fact that the overall computation is quadratic with respect to the sequence length $n$. This may be tolerable in settings like machine translation, where a typical document consists of less than $n<100$ words and a wide network is often used (i.e. $d$ is large); however, for reading comprehension tasks, where long documents and narrow networks are typical (i.e. $n \gg d$), self-attention becomes expensive. In addition, bidirectional recurrent units have the intrinsic problem that their sequential nature precludes parallel processing. 

Admittedly, dilation has its limitations. It requires more overall computations than recurrent nets and the reception region is predetermined. We argue that to provide answers as a web service, one cares more about the response latency for a single question. Therefore, a short compute of the longest computation is favored. 

\subsection{Baseline Models: BiDAF and DrQA}
\begin{figure}
	\centerline{
    \includegraphics[width=\textwidth]{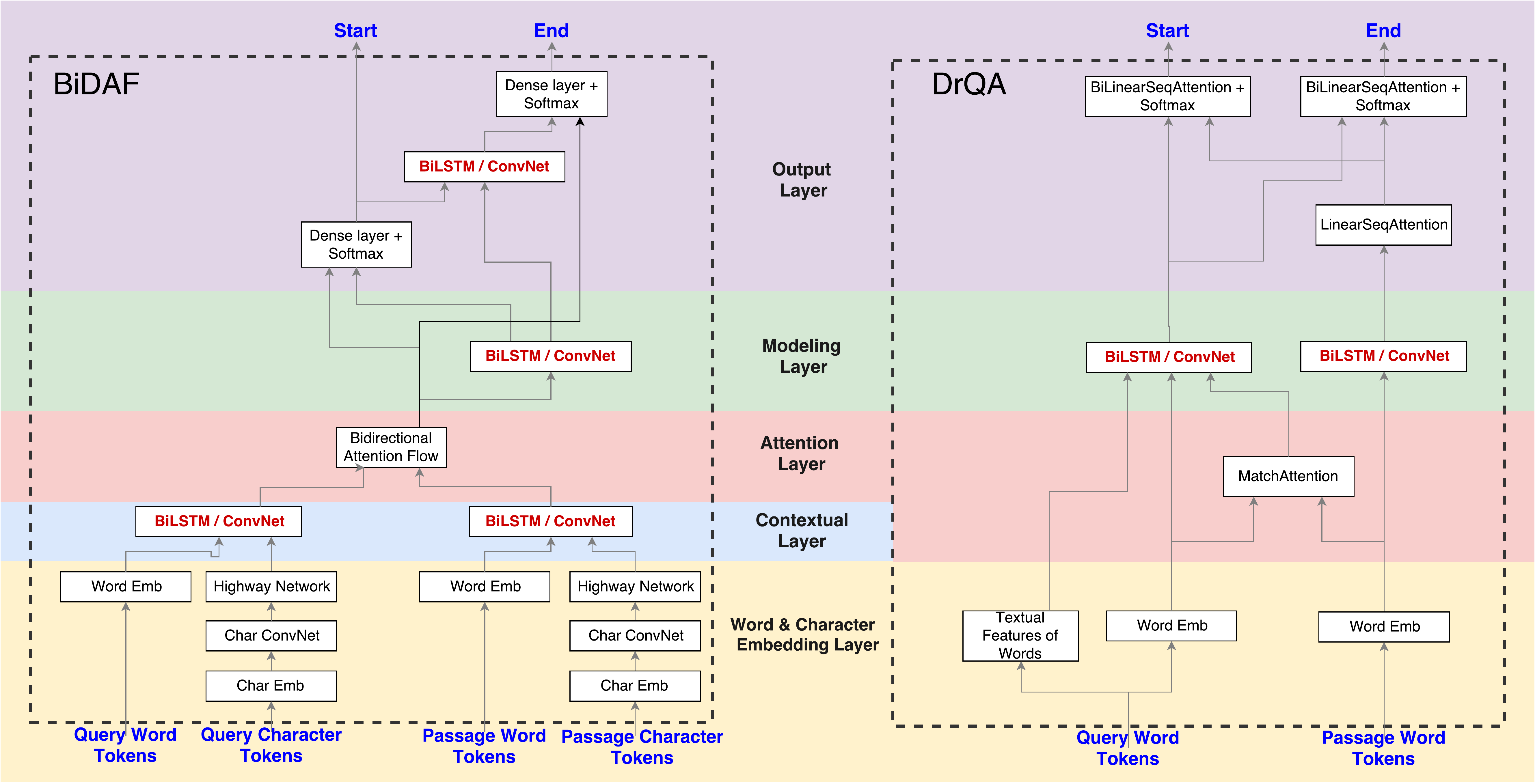}
}
    \caption{Schematic layouts of the BiDAF (\emph{left}) and DrQA (\emph{right}) architectures. We propose to replace all occurrences of BiLSTMs with diluted ConvNet structures.}
    \label{fig:bidaf_arch}    \label{fig:drqa_arch}
\end{figure}

Now we briefly describe 
two popular open-sourced question answering systems: Bi-directional Attention Flow (BiDAF)~\citep{seo2016bidirectional} and DrQA~\citep{chen2017reading}, which are relevant to our study.
Figure~\ref{fig:bidaf_arch} shows schematic layouts of their respective model structures and highlight the BiLSTM layers (in red) which are the bottleneck for inference speed.
Both models require LSTMs to encode the query and passage. 
BiDAF is more complex than DrQA, with two more LSTMs to make the classification decision. 


\paragraph{The BiDAF} model~\citep{seo2016bidirectional} introduces a bidirectional attention flow to help passing information between the passage and the query.
It  has six components: 1) character embedding layer, 2) word embedding layer, 3) contextual layer, 4) attention flow layer, 5) modeling layer, and 6) output layer, but only three of them contains LSTMs. 
The \emph{contextual layer} encodes the passage and the query with two bidirectional LSTMs with shared weights. 
The \emph{modeling layer} further employs a two-layer stacked bidirectional LSTM to extract the higher order features of the words in the passage. 
The \emph{output layer} uses yet another bidirectional layers to produces features for predicting the end of the answer span.

\paragraph{The DrQA} system~\citep{chen2017reading} has a document retriever and a document reader. 
The \emph{document retriever} simply uses pre-defined features to retrieve documents when the corresponding passage is not given in the question. 
The \emph{document reader} uses two 3-layer stacked bidirectional LSTMs to encode the query and the passage/document respectively.

\subsection{Related Work}

\paragraph{Reading Comprehension Models.}
After the release of the Stanford Question Answering Dataset \citep{rajpurkar2016squad}, reading comprehension models kept springing up in the past year. 
All of them use recurrent neural networks~\citep{hochreiter1997long, cho2014gru} as a common component, and most of the top performed models uses attention~\citep{bahdanau2014neural} in addition. 
RNet~\citep{wang2017rnet}, demonstrates the effectiveness of the self-attention modules. 
Document Reader (DrQA)~\citep{chen2017reading}  provides an question answering system using a document database. 
\cite{Hu2017ReinforcedMnemonicReader} demonstrated how reinforcement learning can benefits the training procedure. 
SmartNet~\citep{hermann2015smartnet} proposed to mechanism to keep refining the prediction.

\paragraph{ConvNets for Sequence Generation.} 
There has been a lot of effort in applying ConvNet architectures to reduce the sequential computation in sequence to sequence models such as Extended Neural GPU\citep{kaiser2016am}, ByteNet~\citep{ kalchbrenner2016linear} and ConvS2S~\citet{gehring2017convseq}.
In these models, the number of operations required to relate signals from two arbitrary input or output positions grows with the distance between positions, linearly for ConvS2S and logarithmically for ByteNet.
The improvements in speed are limited to these generative models, because the decoding procedure still needs to be done token by token -- and is therefore inherently linear  with respect to the length of the to decoding sequence.
In comparison, there is no generation in reading comprehension models ~\citep{seo2016bidirectional, chen2017reading}, and much more impressive speedups are possible through the application of ConvNets.  





\section{Model Specifics}
\begin{figure}
	\centering
    \includegraphics[width=0.7\textwidth]{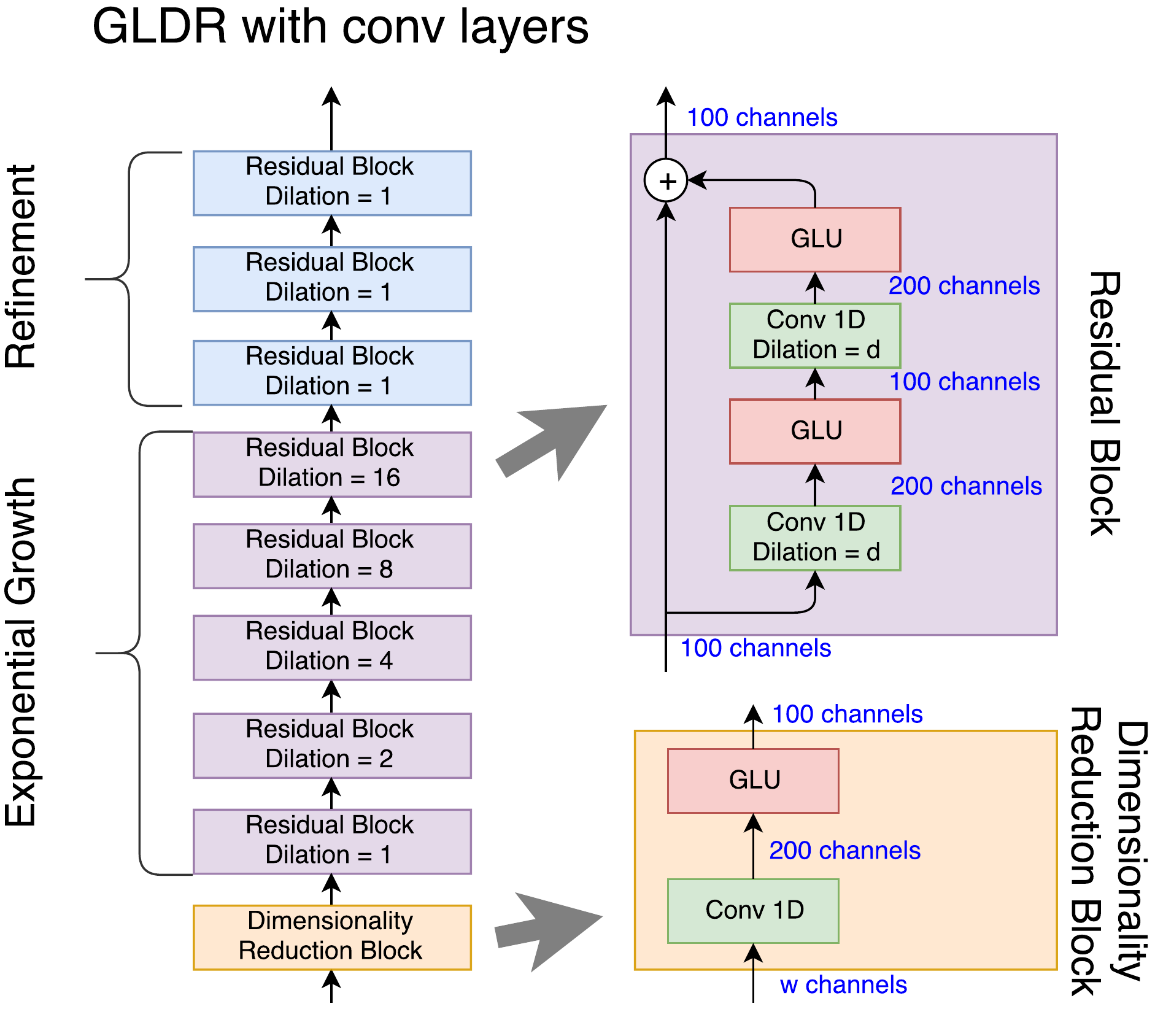}
   \vspace{-3ex}
    \caption{Schematic Layout of the \modelname{}. The \modelshort{} begins with dimensionality reduction to 200 channels through the use of a dimensionality reduction block (\emph{bottom right}). Subsequently a short sequence of residual blocks (\emph{right}) is used to increase the receptive field of the convolution exponentially. The output is further processed with a few layers of standard convolution (dilation 1).}
    \label{fig:block}
\end{figure}

The basic principle behind our approach is simple, yet very effective (as will be demonstrated in our experiments). We substitute the bidirectional sequence models with a simple convolutional network with repeated dilated convolutional layers. The receptive field of this convolutional network grows exponentially with depth and soon encompasses a long sequence, essentially enabling it to capture similar long-term dependencies as an actual sequential model. The compelling advantage of our approach is that the processing time is drastically reduced because convolution can be parallelized across the input passage. In this section we provide some details on our architecture. 
Figure \ref{fig:block} depicts a schematic layout of our proposed model, which we refer to as \emph{\modelname{} (\modelshort{}). } It consists of two basic components, the dimensionality reduction and the residual block. In the following, we explain both in detail and then provide specifics about the application to BiDAF and DrQA.  


\paragraph{The Dimensionality Reduction Block} 
reduces the input (consisting of word embeddings and possibly other features) to a fixed dimensionality of 100 channels. It consists of a normal convolution (kernel size 3) and a gated linear unit (GLU) as activation and we train it with additional  dropout regularization on the input~\citep{hinton2012dropout}.

\paragraph{The Residual Block} is the key ingredient to perform dilated convolution. It has a two-layer ConvNet with GLU activations (optionally with input dropout). The output of this small two-layer ConvNet is later summed up with the input allowing the ConvNet to learning only the residual of the transformation, similar to ResNet~\cite{he16residual}. For simplicity all the convolutions are of kernel size 3 while the dilations vary across layers. To be more specific, the dilations of the convolutions in the first few residual blocks are increased exponentially ($1, 2, 4, 8, \cdots$) with the purpose to increase the receptive field. After a small number of $O(\log(n))$ layers, the receptive field is wide enough and we switch back to 
normal convolutions for further refinement. 


To show ConvNets are not limited to a particular architecture of question answering models, we apply our \modelshort to two popular open-sourced question answering systems: Bi-directional Attention Flow (BiDAF)~\citep{seo2016bidirectional} and DrQA~\citep{chen2017reading}.

\paragraph{Convolutional BiDAF.}
In our convolutional version of BiDAF, we replaced all bidirectional LSTMs with \modelshorts. We have two 5-layer \modelshorts in the contextual layer whose weights are un-tied. In the modeling layer, a 17-layer \modelshort with dilation 1, 2, 4, 8, 16 in the first 5 residual blocks is used, which results in a reception region of 65 words. A 3-layer \modelshort replaces the bidirectional LSTM in the output layer. For simplicity, we use same-padding and kernel size 3 for all convolutions unless specified. The hidden size of all \modelshorts is 100 which is the same as the LSTMs in BiDAF.

\paragraph{Convolutional DrQA.}
Since the query is much shorter than the document, we can afford using a 17-layer \modelshort without dilation for encoding the query. However, a 9-layer \modelshort whose 4 residual blocks have dilation 1, 2, 4, and 8, respectively is used to capture the context information in the passage, which results in a receptive region of 33 words. The hidden size of the ConvNet is 128 which matches that of the LSTMs in DrQA.

\section{Experiments}
We perform experiments on two data sets, the Stanford Question Answering Dataset (SQuAD) \citep{rajpurkar2016squad} and TriviaQA~\citep{joshi2017trivia}. In the following we describe the experimental setup, provide details on both data sets and elaborate on our experimental findings.

\paragraph{Experiment Setup.} We evaluate convolutional versions of BiDAF  and DrQA, which we refer to as Conv-BiDAF and Conv-DrQA respectively. We adopt the open-sourced BiDAF implementation\footnote{\url{https://github.com/allenai/bi-att-flow}} which is written in TensorFlow \citep{abadi2016tensorflow} and DrQA implementation\footnote{\url{https://github.com/facebookresearch/DrQA}} in PyTorch \footnote{\url{http://pytorch.org/}}, and follow their preprocessing and experimental setup. The models are trained with either a NVIDIA Tesla P100 GPU or a NVIDIA Titan X (Pascal) GPU, but the latter is used exclusively for all timing experiments. Across all experiments we only time GPU eclipsed time (i.e. forward and backward passes through the networks) since CPU bounded operations are not our focuses.

\subsection{The Stanford Question Answering Dataset (SQuAD)} 

SQuAD is one of the most popular reading comprehension datasets and contains over 100K questions-answer-passage tuples. The data set was labeled by crowdsource workers who, given a passage, were asked to generate questions based on it. For each passage another group of workers attempted to highlight a span in the passage as the answer. This ensures that the passage also contains sufficient information and the answer is always present. 
We evaluate our experiments on the SQuAD validation set (which is publically available) as the secret test set is guarded with access limitations. 

\paragraph{Experiment Setup}
For BiDAF and Conv BiDAF we try out a few optimization settings and picked the best one based on the validation set. 

For the final BiDAF model, we use a batch size of 60, dropout rate 0.2 \citep{srivastava2014dropout}, and train it for $60000$ iterations. In addition, we use stochastic gradient descend with momentum $0.1$ and weight decay $10^{-4}$, and decay the learning rate by a factor of 10 every 20000 iterations, which improves the performance of BiDAF slightly compared to training it with Adam \citep{kingma2014adam} for 20000 iterations suggested by \cite{seo2016bidirectional}\footnote{Based on \url{https://github.com/allenai/bi-att-flow/issues/10}, the authors switch from Adadelta to Adam.}.

For our Conv BiDAF, we train the model for $60000$ iterations with the Adam optimizer \citep{kingma2014adam} using the default settings in TensorFlow ($\alpha = 0.0001, \beta_1 = 0.9, \beta_2 = 0.999$), drop $\alpha$ by a factor of $10$ every $20000$ iterations, and use an additional word dropout rate $0.1$ \citep{dai2015semi}. Word dropout isn't found helpful for BiDAF in our experiment. Because of the GPU memory constraint, the model is trained with the documents shorter than or equal to $400$ word tokens as what the authors did in the paper. 

For all DrQA variants, we adopt batch size 32, dropout rate 0.3, and train both models for $60$ epochs with Adamax \citep{kingma2014adam} optimizer using the default setting in PyTorch  ($\alpha = 0.002, \beta_1 = 0.9, \beta_2 = 0.999$). Weight decay and word dropout doesn't result in a fair amount of improvement on either of the models, so they are abandoned in the reported models. The models are trained on the SQuAD training set without removing long documents.

\begin{figure}[t]
    \centering
    \includegraphics[width=1\textwidth]{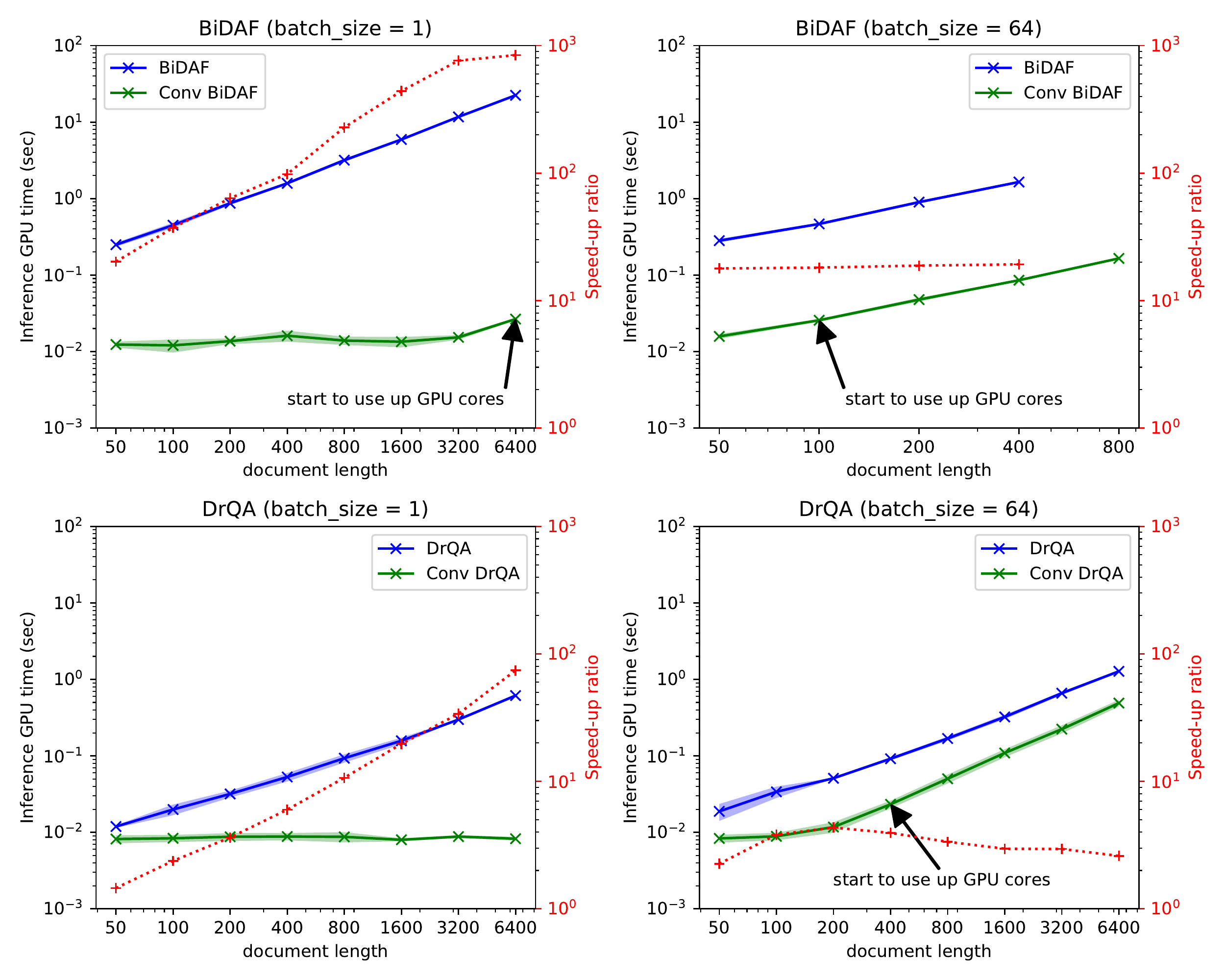}
    \caption{Inference GPU time of four models with batch size $1$ or $64$. The time spent on data pre-processing and decoding on CPUs are not included. The lines of Convolutional models should be almost flat as long as their are enough cores in the GPU. The DrQA uses the CuDNN LSTM implementation which makes it much faster than BiDAF. The missing points for BiDAF and Conv-BiDAF were caused by running of out GPU memories. We use a single NVIDIA Titan X (Pascal) with 12GB memory is this experiment.
}
    \label{fig:speedup_doc_length}
\end{figure}

\begin{figure}[h]
    \centering
    \includegraphics[width=1\textwidth]{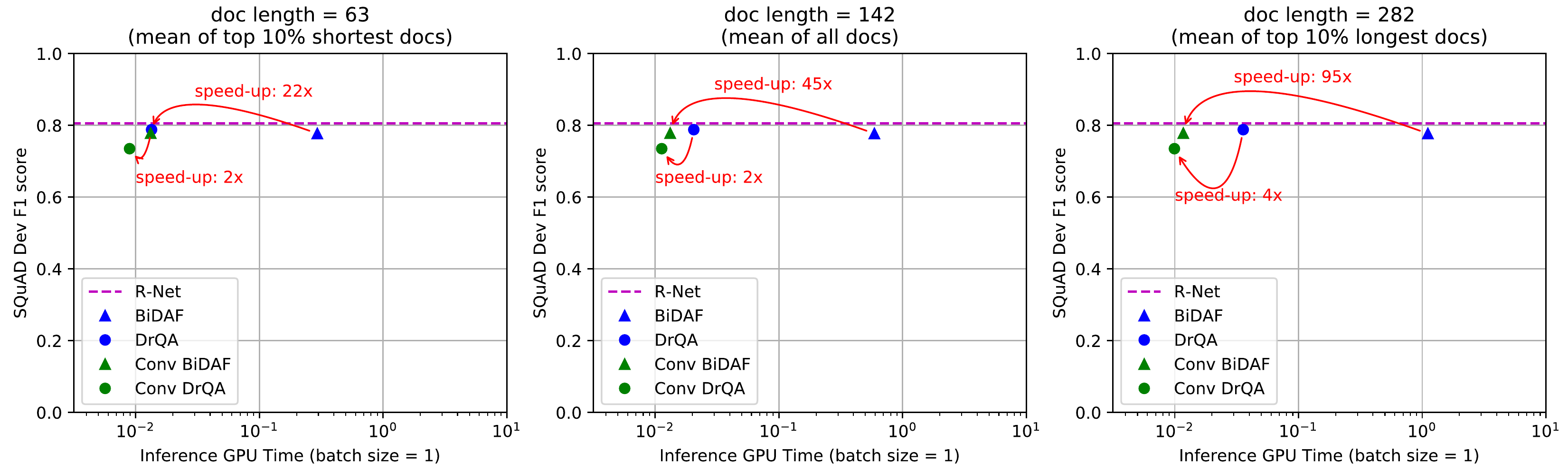}
    \vspace{-3ex}
    \caption{Dev F1 Score on SQuAD vs Inference GPU Time for a variety of doc lengths. Our proposed methods, Conv BiDAF and Conv DrQA, approximately maintain the high accuracies of the original models while introducing tremendous speedups at inference time.}    
    \label{fig:squad_f1_vs_time}
\end{figure}

\paragraph{Results.} \autoref{fig:squad_f1_vs_time} shows the F1 score on the development partition of the SQuAD dataset for the various algorithms, Conv BiDAF, Conv DrQA, BiDAF, DrQA, and R-NET~\citep{wang2017rnet} as a function of inference GPU time. 
The figure shows the results across all documents (middle) and for the top 10\% shortest and longest documents  (left and right plot respectively). Throughout, Conv-BiDAF achieves one to two order of magnitude speed-up at inference time  and performs almost as well as the original BiDAF.The plot also shows the F1-score of the R-Net model for which we have no inference timing (therefore shown as a horizontal line). More detailed comparisons between BiDAF and Conv-BiDAF are shown in \autoref{table:bidaf-vs-cbidaf}.

On \autoref{table:squad_performance}, we show different variants of BiDAF models and their performance.

\begin{table}[h]
\centering
\begin{tabular}{l|l|l}
Model         & BiDAF   & Conv BiDAF (5-17-3) \\ 
\# of params      & 2.70M   & 2.76M                     \\ 
\hline
Dev EM    & $67.66 \pm 0.41$   & $68.28 \pm 0.56$                     \\ 
Dev F1    & $77.29 \pm 0.38$   & $77.27 \pm 0.41$         \\ 
\hline
Training GPU time (h) until convergence & $21.5$\footnotemark\label{fnm:traintime}  & $3.4$ (6.3x)                \\ 
Training GPU time (sec) per iteration, batch size = 60  & $3.88 \pm 2.14$  & $ 0.20 \pm 0.14$ (19x)           \\ 
\hline
Inference GPU time (sec) per iteration, batch size = 60 & $1.74 \pm 0.75 $ & $ 0.081 \pm 0.002 $ (21x)         \\ 
Inference GPU time (sec) per iteration, batch size = 1  & $1.58 \pm 0.06 $ & $ 0.016 \pm 0.003$ (98x)   \\
\end{tabular}
\caption{BiDAF v.s. Conv BiDAF. 
EM stands for exact match score.}
\label{table:bidaf-vs-cbidaf}
\end{table}
\footnotetext{To make it fair we report the training time of BiDAF for 20000 iterations. Both our BiDAF and Conv-BiDAF were trained for 60000 iterations to reach the best dev F1 based on hyperparameter tuning, but \cite{seo2016bidirectional} suggests training it for only 20000 iterations which achieves a sightly worse results for BiDAF.}

\begin{table}[h]
\centering
\begin{tabular}{l|r|r|r}
Model                      & \# of params & Dev EM    & Dev F1    \\ \hline
BiDAF (trained by us)        & 2.70M        & 67.94     & 77.65 \\
Conv BiDAF (5-17-3 conv layers)  &  2.76M       & 68.87     & 77.76 \\
Conv BiDAF (9-9-3 conv layers)   &  2.76M       & 67.79     & 77.11 \\
Conv BiDAF (0-31-3 conv layers)  &  3.36M       & 63.52     & 72.39 \\
Conv BiDAF (11-51-3 conv layers) &  5.53M       & \textbf{69.49}     & \textbf{78.15} \\ 
Conv BiDAF (31-31-3 conv layers) &  6.73M       & 68.69     & 77.61 \\
\hline
DrQA (trained by us)             & 33.82M       & \textbf{69.85}& \textbf{78.96} \\
Conv DrQA (9 conv layers)        & 32.95M       & 62.65     & 73.35 \\
\end{tabular}
\caption{Comparing variants with different number of layers. EM stands for exact match score. The scores are multiplied by $100$. DrQA uses a much larger pre-trained word embedding resulting in more parameters.}
\label{table:squad_performance}
\end{table}

\begin{table}[h]
\centering
\begin{tabular}{l|r|r|r}
Model                            & Dev EM    & Dev F1    \\ \hline
Conv BiDAF (5-17-3 conv layers)  & \textbf{68.87}     & \textbf{77.76} \\
without dilation                 & 68.15     & 76.99 \\
Using ReLU instead of GLU        & 63.91     & 73.39 \\
\end{tabular}
\caption{Ablation Test. EM stands for exact match score. The scores are multiplied by $100$. Without residual learning (shortcut connection) the model cannot learning meaningful information and diverges.}
\label{table:ablation}
\end{table}

\autoref{table:ablation}   compares various structural choices for the convolutional architecture. BiDAF(11-51-3) produces the best answer quality, but the model is also larger, and therefore more expensive for inference.
Overall, all ConvNet models are much smaller than the LSTM based models, while still able to produce comparable answer quality.

\autoref{table:squad_performance} gives the result of ablation studies. We can see that dilation helps improve answer quality a bit.
The choice of non-linear unit has a huge impact on answer quality with GLU performs much better than ReLU.
Without residual learning (removing the shortcut connection), the model can never converge to fair point.

\subsection{TriviaQA} 

TriviaQA is a large-scale reading comprehension dataset with 95K question-answer pairs and 650K question-answer-evidence tuples which is more challenging than SQuAD because it 1) contains more complex questions, 2) has substantial syntactic and lexical variability in the text, 3) requires a significant amount of cross-sentence reasoning, and 4) the answer and the sufficient information are guaranteed in the evidence. For each question-answer pair, it used distant supervision to provide relevant evidence from wikipedia or web search. Besides the full development and test set. A verified subset for each is also provided. 

With the same hyperparamters used for SQuAD, our Conv DrQA outperforms all models reported in the published literatures while being slightly worse than the unpublished ones on the wiki split leader-board. Again, we can see a trade-off between performance and speed.

\paragraph{Experiment Setup}
We process the data into SQuAD format with the script provided by Trivia QA\footnote{https://github.com/mandarjoshi90/triviaqa}. Precisely, for each document in the candidate set of a question-answer pair, it produces a question-answer-evidence tuple for training as long as any of the answers appear in the first 800 tokens in the document.  For evaluation, we truncate each document down to 1600 tokens and predict a span among them. We follow the suggestion from \cite{joshi2017trivia} to use only the first 80K question-answer-evidence tuples (out of 529K) of the Web split of TriviaQA for training.

\paragraph{Results.}
On Wikipedia split of TriviaQA, our proposed Conv DrQA is slight worse than our DrQA baseline which beats all previous models, it can still be on a par with the previous state-of-the-art performance of recurrent networks. The numbers are shown in \autoref{table:trivia_results} The Conv DrQA model only encode every 33 tokens in the passage, which shows that such a small context is enough most of the question.

\begin{figure}[h]
    \centering
    \includegraphics[width=1\textwidth]{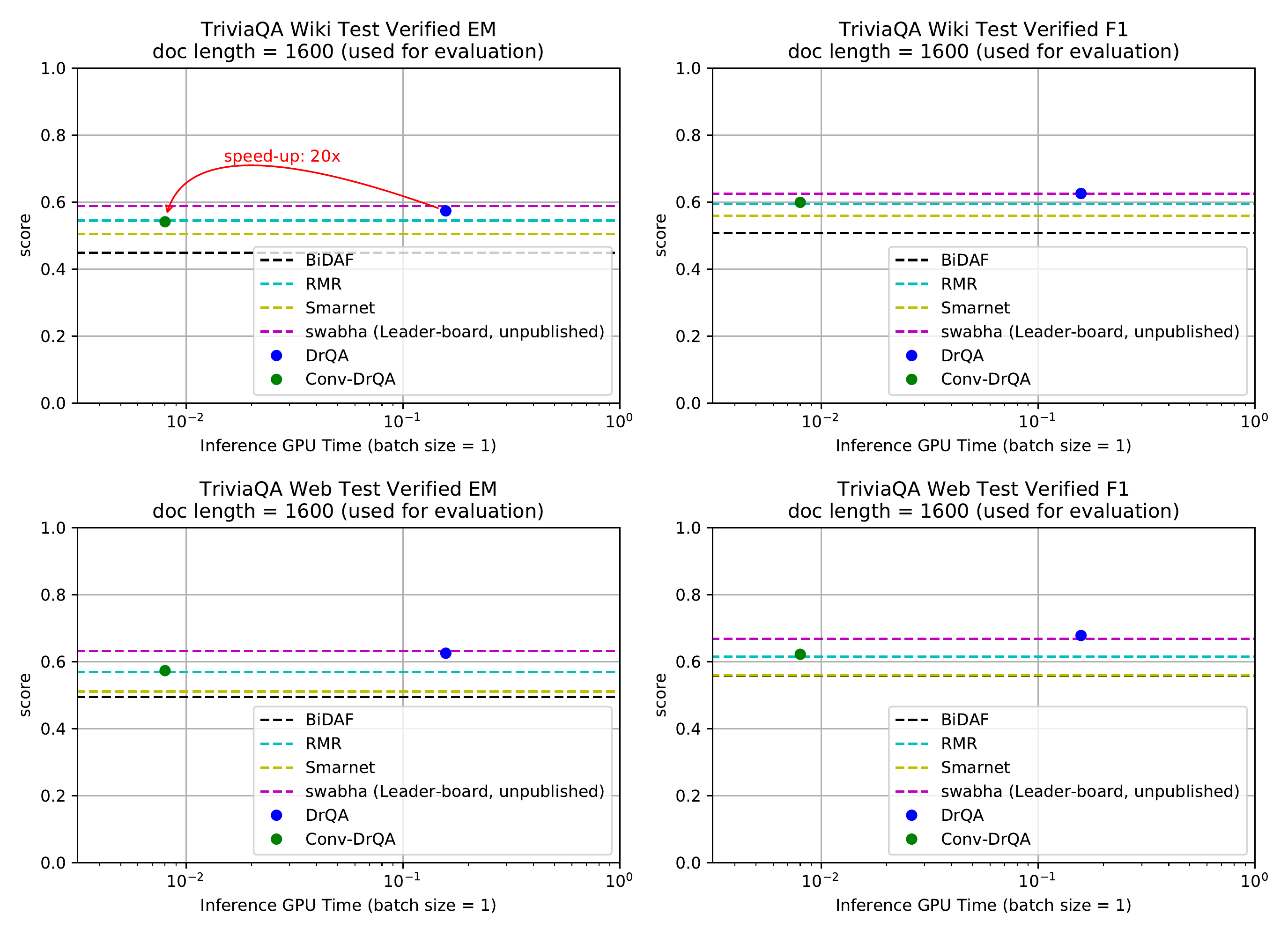}
    \vspace{-3ex}
    \caption{Test exact match and F1 Score on Wikipeida and web split of TriviaQA vs Inference GPU Time for the maximum document length used in our evaluation. Our proposed methods, Conv BiDAF and Conv DrQA, approximately maintain the high accuracies of the original models while introducing tremendous speedups at inference time.}    
    \label{fig:trivia_f1_vs_time}
\end{figure}

\subsection{Memory usage} 
\begin{figure}
	\centering
    \includegraphics[width=0.6\textwidth]{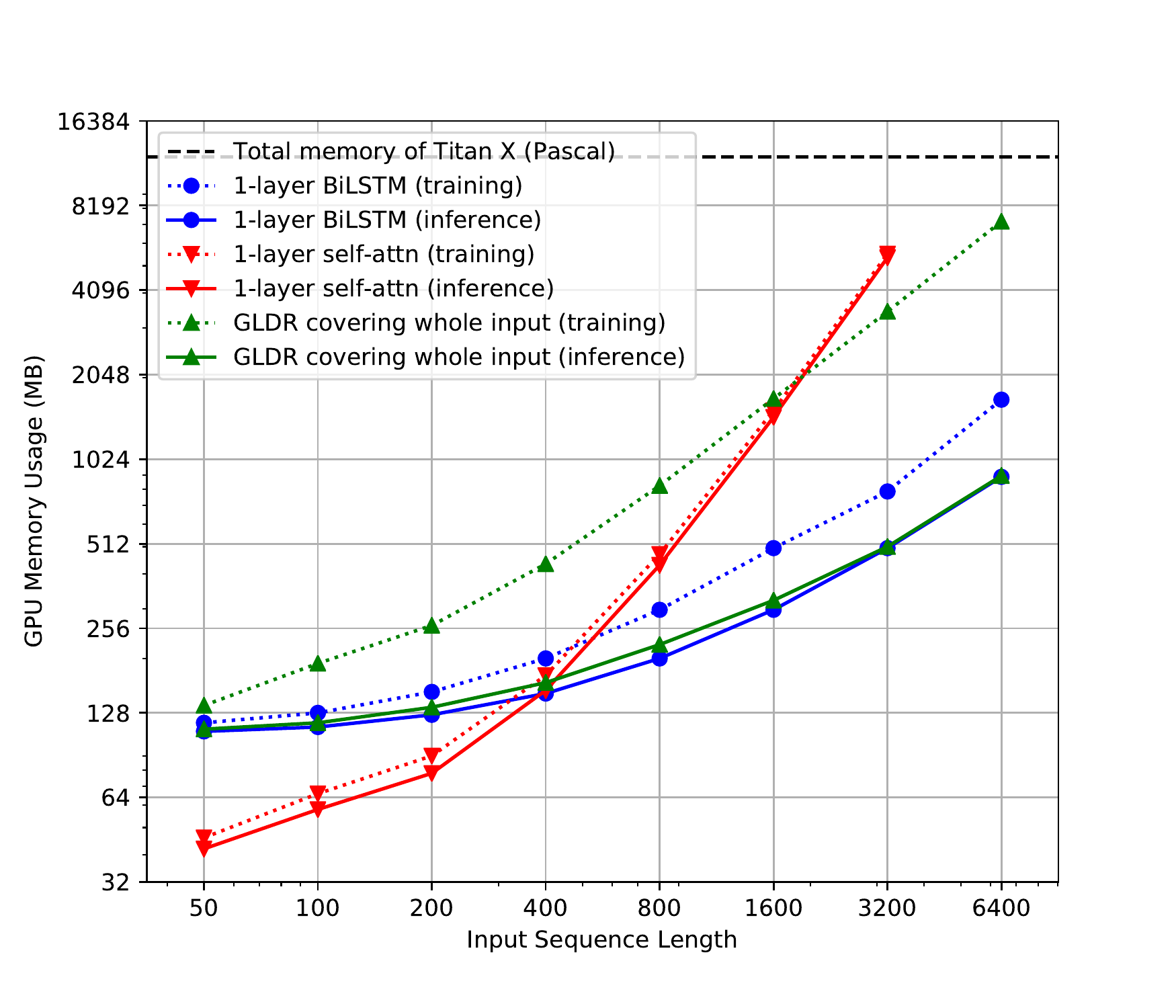}
   \vspace{-3ex}
    \caption{GPU memory usage of a single layer self-attention versus a GLDR with enough depth whose receptive field covers the whole sequence. The number of convolution layers of GLDR is 15 when the input sequence length is 50, and increases by 2 (adding one more residual block) each time the sequence length doubles. The missing points are caused by out of GPU memory.}
    \label{fig:memory}
\end{figure}

\autoref{fig:memory} compares the memory usage of different types of models.
We can see that the quadratic growth of self-attention may prevents its use with most real-world documents.
The experiment is done as follows. We fix input size and the hidden size of the layers to be 100 and use the PyTorch implementation. The batch size is fixed to 64. For self-attention the attention matrix is computed by forwarding the input vectors through a dense layer with ReLU activation and then taking the inner product between any pairs of two outputs of this dense layer.


\section{Conclusion}
%
We propose a convolutional architecture as an alternative to the  recurrent architectures typically used in reading comprehension models. %
By using simple dilated convolutional units in place of recurrent models, we achieve results comparable to the state of the art on two question answering tasks, while at the same time achieving up to two orders of magnitude speedups at inference time. Most applications of question answering (e.g. search engines or mobile assistant) are particularly sensitive to latency  and even a small speedup can be a huge improvement. 

Our results raise the question for which other tasks in NLP sequence models are not necessary and can be replaced by dilated convolution.  If the purpose of the sequence model is to enable long-range dependencies,  it may be that convolution with very large receptive fields could be sufficient. In this paper, we provide evidence that in the case of reading comprehension this may be the case.

\subsubsection*{Acknowledgments}
We are grateful to Avinash Atreya, Neil Houlsby, Tom Kwiatkowski, and Lukasz Kaiser for helpful discussions during the course of this work.


\bibliography{sections/refs}
\bibliographystyle{iclr2018_conference}

\newpage
\appendix
\section{More Experiments}
\subsection{TriviaQA scores}
We report the full and verified exact-match and F1 scores in \autoref{table:trivia_results}

\begin{table}[h]
\centering
\begin{tabular}{l| l|r|r|r|r}
& & \multicolumn{2}{c}{Full} & \multicolumn{2}{c}{Verified} \\
Dataset &Model  &  EM &  F1 & EM &  F1 \\ 
\hline
\multirow{7}{*}{Wikipedia} 
&BiDAF \citep{joshi2017trivia} & 40.32 & 45.91 & 44.86 & 50.71 \\
&RMR \citep{Hu2017ReinforcedMnemonicReader} & 46.94 & 52.85 & 54.45 & 59.46  \\
&Smarnet \citep{Chen2017SmarnetTM} & 42.41 & 48.84 & 50.51 & 55.90  \\
&swabha\_ankur\_tom (unpublished) (Leader-board \cref{fn:leaderboard}) & 51.59 & 55.95 & 58.90 & 62.53 \\
&DocQA \citep{clark2017docqa} & - & - & - & -  \\
&DrQA & 52.58 & 58.17 & 57.36 & 62.55 \\ 
&Conv DrQA & 49.01 & 54.52 & 54.11 & 59.90 \\ 
\hline
\multirow{6}{*}{Web}
&BiDAF \citep{joshi2017trivia} & 40.74 & 47.06 & 49.54 & 55.80 \\
&RMR \citep{Hu2017ReinforcedMnemonicReader} & 46.65 & 52.89 & 56.96 & 61.48 \\
&Smarnet \citep{Chen2017SmarnetTM} & 40.87 & 47.09 & 51.11 & 55.98  \\
&swabha\_ankur\_tom (unpublished)  & 53.75 & 58.57 & 63.20 & 66.88 \\
&DocQA \citep{clark2017docqa} (Leader-board \cref{fn:leaderboard}) & 66.37* & 71.32* & 79.97* & 83.70*  \\
&DrQA  & 51.49 & 57.87 & 62.55 & 67.84 \\ 
& Conv DrQA  & 47.77 & 54.33 &57.35 & 62.23 \\ 
\end{tabular}
\caption{TriviaQA Performance. The scores are multiplied by $100$. $^*$Doc QA \citep{clark2017docqa} uses a different training strategy on the whole training set (containing 529K question-answer-evidence tuples), while we follow \cite{joshi2017trivia} training on only the first 80K question-answer-evidence tuples in the training set.
}
\label{table:trivia_results}
\end{table}

\footnotetext{\label{fn:leaderboard} The best scores recorded on Nov 9, 2017 from the Leader-board \url{https://competitions.codalab.org/competitions/17208\#results}.}

\subsection{Open-Domain Question Answering}
\nl{
Why do we want to follow \cite{chen2017reading}?
How is retrieval approach related to our goal of replacing LSTMs with ConvNets?}
\fw{The retrieval part is not related to our approach, but these are other datasets that show that Conv DrQA is as good as DrQA.}

\nl{why ConvDrQA produces worse quality on triviaQA, but better quality on open QA tasks?} \fw{I guess it might be that since the accuracy is quite low, it may just answers the questions that don't require long dependency, but we have to verify it.}

\nl{is ConvDrQA faster on these tasks compared to DrQA?} \fw{Yes, it is.}

\fw{looks like we should just keep the results here as more experiments to avoid misleading the readers?}

\begin{table}[]
\centering
\begin{tabular}{l|rr|rr}
             & \multicolumn{2}{l|}{\# of questions} & \multicolumn{2}{l}{\begin{tabular}[c]{@{}l@{}}\# of tuples for\\ Distant Supervision\end{tabular}} \\ \cline{2-5} 
Dataset      & Full Train           & Test          & Train                                            & Dev                                              \\ \hline
CuratedTrec  & 1,486                 & 694           & 2,935                                             & 735                                              \\ 
WebQuestions & 3,778                 & 2,032          & 5,736                                             & 1,464                                             \\ 
WikiMovies   & 96,185                & 9,952          & 75,174                                            & 18,752                                           
\end{tabular}
\caption{Statistics of the open-domain datasets.}
\label{table:odqa}
\end{table}

Following \cite{chen2017reading}, we further evaluate the full Conv DrQA system which uses the document retriever in DrQA to filter the documents and answers open-domain questions. 
We test the model on three open-domain query sets -- CurateTREC \cite{Baudis:2015:MQA:2960928.2960935}, WebQuestions \cite{DBLP:conf/emnlp/BerantCFL13}, and WikiMovies \cite{DBLP:conf/emnlp/MillerFDKBW16}. 
\autoref{table:odqa} compares these datasets.
We show that Conv DrQA achieves comparable results as DrQA on these open-domain questions.

\paragraph{Experiment Setup}
On open-domain datasets, only question-answer pairs are provided.
We use the Wikipedia pages provided by \cite{chen2017reading} which contains 5 million pages to provide distant supervision.
To be more specific, given a question-answer pair, we find the Wikipedia pages containing the answer and treat it as an evidence to generate a question-answer-evidence tuple.
However, this distant supervision can only be used for generating training data.
To evaluate the model, we have use the document retriever to retrieve relevant documents. The document reader rank all the Wikipedia pages using TF-IDF and bigram-hash. The document reader in DrQA or Conv DrQA finds an answer span in the top 5 documents.
We start with a document reader pre-trained on SQuAD and fine-tune it with distant supervision as proposed by \cite{chen2017reading}.
We randomly sample 20\% of the question-answer-evidence tuples in the original training set as the development set. We tune the the learning rate and the dropout rate, and select our model based on the exact match score on this development set.

We use the DrQA source code to generate the distant supervision data; however, we observe different statistics from what was reported by \cite{chen2017reading}. The statistics of the generated distant supervision data is shown in \autoref{table:odqa}.

\paragraph{Results}
\autoref{table:finetuning_results} shows the test exact match of the fine-tuning models. Our Conv DrQA outperforms DrQA on the CuratedTrec and the WikiMovies datasets, while being a little worse on the WebQuestions dataset.

\begin{table}[]
\centering
\begin{tabular}{l|lll}
                        & CuratedTrec & WebQuestions & WikiMovies \\ \hline
DrQA \cite{chen2017reading}          & 25.7         & 19.5         & 34.3       \\
DrQA (reproduced by us) & 25.9         & \textbf{20.1}         & 35.5       \\
Conv DrQA               & \textbf{27.2 }        & 19.1         & \textbf{36.2}      
\end{tabular}

\caption{Fine-tuning results on open-domain datasets. All the scores are the exact match scores on the test set and multiplied by 100.}
\label{table:finetuning_results}
\end{table}

\end{document}